\pdfoutput=1

\documentclass[11pt]{article}

\usepackage[final]{acl}
\usepackage{times}
\usepackage{latexsym}
\usepackage[T1]{fontenc}
\usepackage[utf8]{inputenc}
\usepackage{xcolor}
\usepackage{tcolorbox}
\usepackage{microtype}
\usepackage{inconsolata}
\usepackage{graphicx}
\usepackage{subcaption}
\usepackage{booktabs}
\usepackage{multirow}
\usepackage{tablefootnote}
\usepackage{hyperref}
\usepackage{float}
\usepackage{colortbl}
\usepackage{subcaption}
\usepackage{multicol}
\usepackage{amsmath}
\usepackage{cleveref}
\definecolor{myblue}{RGB}{150, 150, 230}

\usepackage{amsmath,amsfonts,bm}

\def\eqref#1{equation~\ref{#1}}

\def\1{\bm{1}}

\DeclareMathAlphabet{\mathsfit}{\encodingdefault}{\sfdefault}{m}{sl}
\SetMathAlphabet{\mathsfit}{bold}{\encodingdefault}{\sfdefault}{bx}{n}

\def\sD{{\mathbb{D}}}

\DeclareMathOperator*{\VLM}{LVLM}
\DeclareMathOperator*{\DPO}{DPO}
\DeclareMathOperator*{\SFT}{SFT}

\title{Self-Rewarding Large Vision-Language Models for Optimizing Prompts \\in Text-to-Image Generation}

\author{Hongji Yang, 
Yucheng Zhou,
Wencheng Han,
Jianbing Shen\footnotemark[1]\\\\
SKL-IOTSC, CIS, University of Macau \\ 
{\tt\small yc47942@um.edu.mo, yucheng.zhou@connect.um.edu.mo, wencheng256@gmail.com}
}

\begin{document}
\maketitle

\renewcommand{\thefootnote}{\fnsymbol{footnote}} 
\footnotetext[1]{Corresponding author. This work was supported by the Science and Technology Development Fund of Macau SAR (FDCT) under grants 0102/2023/RIA2 and 0154/2022/A3 and 001/2024/SKL and CG2025-IOTSC, the National Natural Science Foundation of China (No. 624B2002), and the Jiangyin Hi-tech Industrial DevelopmentZone under the Taihu Innovation Scheme (EF2025-00003-SKL-IOTSC).}

\begin{abstract}
Text-to-image models are powerful for producing high-quality images based on given text prompts, but crafting these prompts often requires specialized vocabulary. 
To address this, existing methods train rewriting models with supervision from large amounts of manually annotated data and trained aesthetic assessment models. 
To alleviate the dependence on data scale for model training and the biases introduced by trained models, we propose a novel prompt optimization framework, designed to rephrase a simple user prompt into a sophisticated prompt to a text-to-image model. 
Specifically, we employ the large vision language models (LVLMs) as the solver to rewrite the user prompt, and concurrently, employ LVLMs as a reward model to score the aesthetics and alignment of the images generated by the optimized prompt. 
Instead of laborious human feedback, we exploit the prior knowledge of the LVLM to provide rewards, i.e., AI feedback. Simultaneously, the solver and the reward model are unified into one model and iterated in reinforcement learning to achieve self-improvement by giving a solution and judging itself. 
Results on two popular datasets demonstrate that our method outperforms other strong competitors. 
\end{abstract}
\section{Introduction}
\label{sec:intro}

\begin{figure}[!t]
\centering
    \includegraphics[width=1\linewidth]{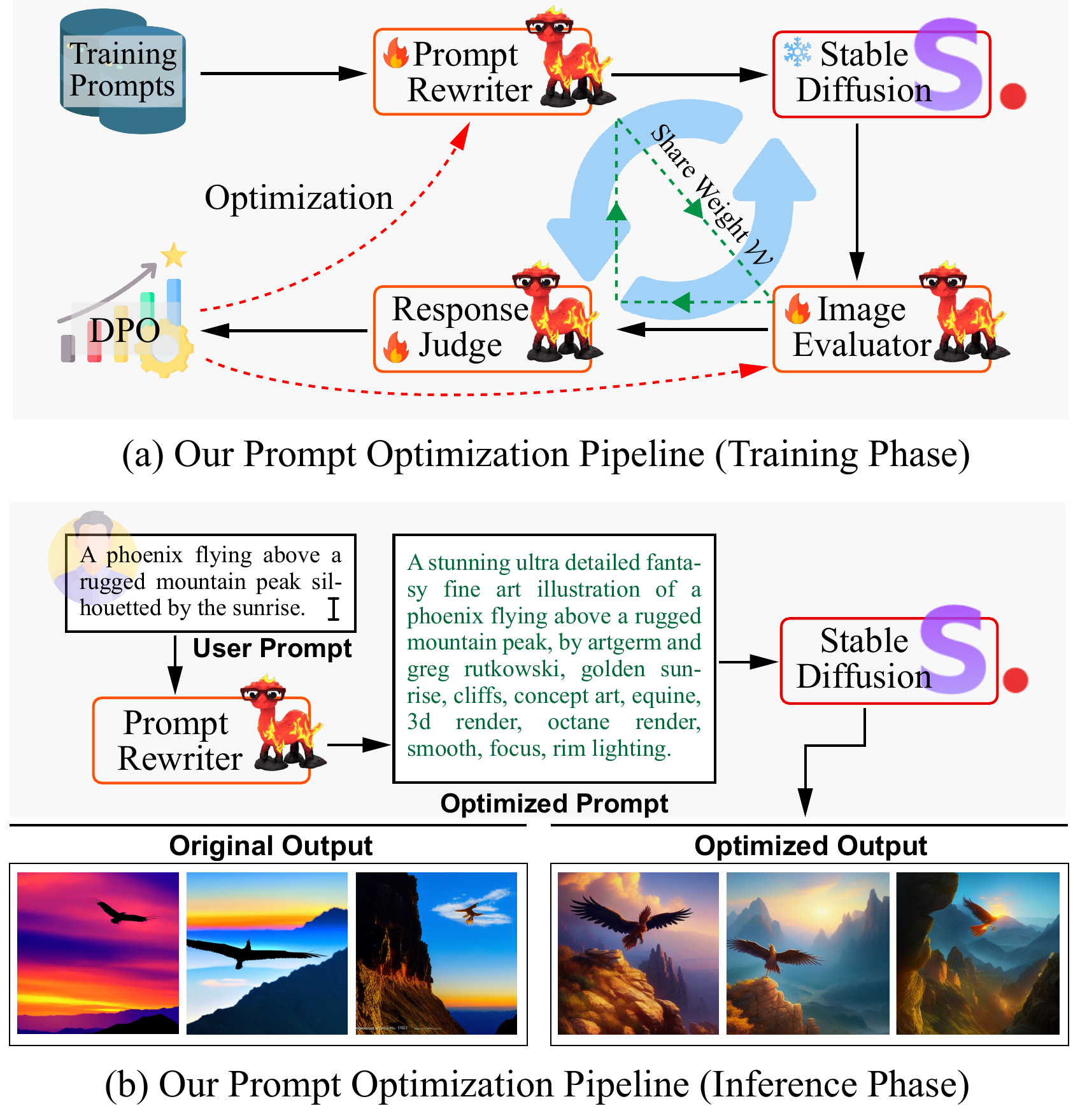}
    \vspace{-8mm}
    \caption{\small The motivation of our prompt optimization pipeline. When training, the large model continuously improves prompt rewriting and image quality evaluation capabilities through self-play without any external sources or models. The generated images from the user prompts and the rewritten prompts. It can be observed that the image from the modified prompt has higher aesthetics.}
    \vspace{-5mm}
    \label{fig1:prompt-image}
\end{figure}

Text-to-image models~\cite{rombach2022high, saharia2022photorealistic, yu2022scaling} can generate diverse high-quality images based on user-provided prompts. 
However, effective interaction with these models requires users to possess specific expertise, including familiarity with specialized vocabularies, e.g., ``\textit{35mm}'' for camera parameters and ``\textit{art by Greg Rutkowski}'' to invoke a particular artistic style. 
As shown in \Cref{fig1:prompt-image}, rewriting prompts according to model-specific knowledge significantly improves the quality of the generated images.

To bridge the gap between laymen and experts in using text-to-image models, some methods \cite{liu2022design, datta2023prompt, oppenlaender2023taxonomy} employ large-scale human-annotated datasets to train rewriting models that produce more professional and effective prompts. 
However, these approaches are prohibitively expensive.
To reduce reliance on high-quality human annotations, some efforts \cite{hao2024optimizing, rosenman2023neuroprompts, cao2023beautifulprompt} leverage specific calculated metrics, i.e., aesthetics and alignment, which are then regarded as rewarding for reinforcement learning (RL).
However, collecting high-quality human-annotated datasets for reward model training is both time-consuming and costly.
Large Vision-Language Models (LVLMs) have recently shown remarkable vision reasoning capabilities. 
Consequently, some studies~\cite{li2022vision, liu2024visual, chen2024mllm,yang2023dawn,yu2023mm} employ LVLMs as evaluators to assess human preferences. 
These methods effectively deliver interpretable AI feedback, offering a more efficient alternative to the time-consuming and labor-intensive human feedback \cite{yuan2024rrhf}.

Besides, previous prompt optimization methods using RL suffer from two limitations: 
(1) They require extensive training data to train an image reward model; 
(2) The reward model remains fixed during the proxy model training, preventing it from learning and improving alongside the proxy model. 
This limitation results in a lack of dynamic feedback throughout the training process.
To mitigate the data limitation and explore dynamic feedback, \citet{yuan2024self} introduce a self-rewarding training strategy, enabling the model to train effectively with limited data while approximating the upper performance bound. It fosters self-improvement or self-play \cite{chen2024self}, wherein the solver generates its judgments or rewards by a continuous iterative DPO \cite{xu2023some}.

In this study, we introduce a self-rewarding prompt optimization framework for text-to-image models. 
This framework leverages an LVLM which functions both as a solver and an evaluator. 
The training pipeline is structured into five key components: model initialization, prompt generation, image generation, LVLM rewarding, and RL training. 
The pipeline is as follows: 
(1) Model initialization: we train the LVLM for prompt optimization on human-annotated prompt rewriting pairs and limited evaluation data, encapsulating LVLM's basic capability to rewrite prompts and assess the preference of generated images. 
(2) Prompt Generation: we employ a large version of LVLM to generate responses for a combination of an existing image quality evaluation dataset with the image evaluation prompt we used. 
(3) Image Generation: the model rewrites the raw prompt according to the instruction and samples multiple results, i.e., ``candidates'', and a fixed text-to-image model is used to generate the corresponding image.
(4) LVLM Rewarding: the LVLM is employed to evaluate the aesthetics and alignment of the image with the original prompt, and is utilized as a rating system to score the images due to endowing with the capability to assess preferences. 
(5) RL Training: we select the highest and lowest-scored candidates to form preference pairs, which are used to train the model and adjust its output preferences by DPO training. 
To further enhance the model's ability to judge image quality, we also make the model itself act as a judge on the model's responses to image aesthetics or alignment to pick the most and least confident responses to construct preference pairs.

The main contributions are summarized below:
\begin{itemize}
    \item \vspace{-2mm}We provide an AI-feedback approach to achieve aesthetic and alignment understanding of images using an LVLM. This approach effectively transforms the LVLM into a reward model to facilitate prompt optimization.
    \item \vspace{-2mm}We introduce self-rewarding training into prompt optimizing for the first time, which obtained prompts with higher quality by iterating the model on a small amount of training data, alleviating the shortcomings of models that require larger and higher quality data for reinforcement learning training.
    \item \vspace{-2mm}In the experiments, we compare our method and other strong competitors on two popular text-to-image datasets, i.e., beautiful-prompt and DiffusionDB. Results show our method achieves state-of-the-art performance.
\end{itemize}

\section{Background}\label{sec:formatting}
\vspace{-1mm}
\subsection{Prompt Rewriting}
\vspace{-1mm}
The purpose of the prompt rewriting is to unlock the maximum potential of text-to-image models. Given an original user prompt $\textbf{x}$, the prompt rewrite models with weight $\theta$ can produce more professional prompts $\textbf{y}$ to help text-to-image models achieve images with more aesthetic pleasure and relevance. The process can be expressed as follows:
\vspace{-1mm}
\begin{equation}
    \begin{aligned}
        \textbf{y} = p(y|\textbf{x};\theta) 
    \end{aligned}
\end{equation}
\vspace{-2mm}

\begin{figure*}[!t]
\centering
\vspace{-2mm}
\includegraphics[width=1\linewidth]{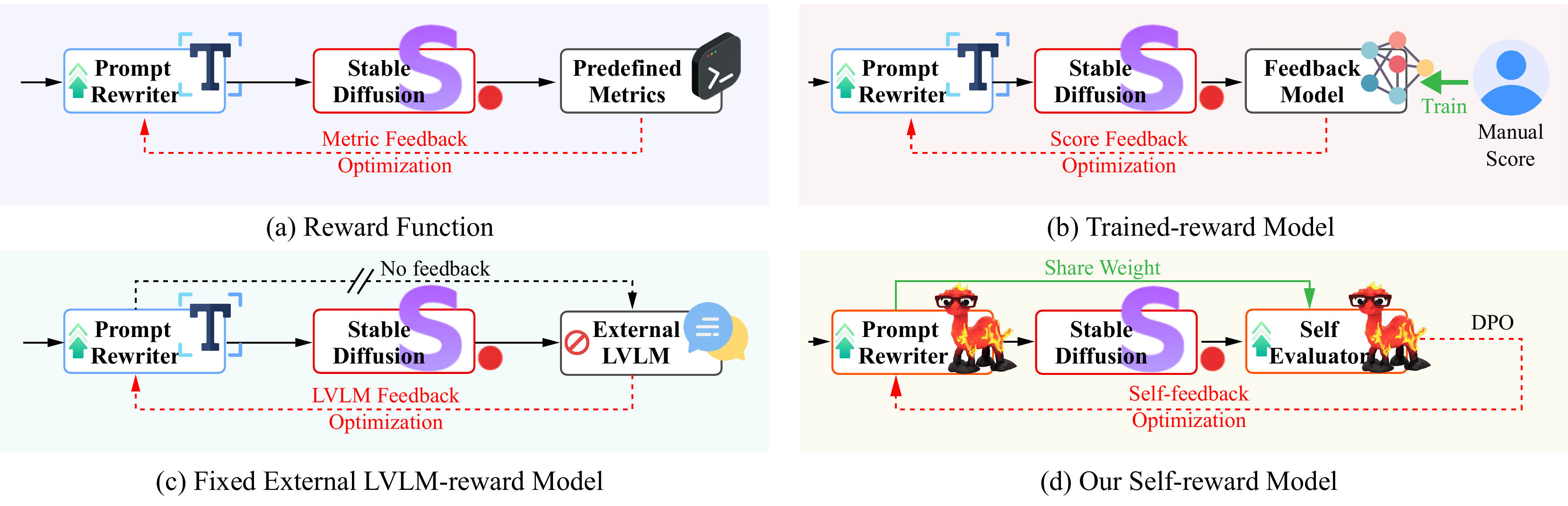}
\vspace{-8mm}
\caption{\small \textbf{Four types of framework in prompt optimizing}. The main difference among them is the reward generation. \textbf{(a)} The reward function (or Metrics) is pre-defined for reinforcement learning, which typically involves mathematical equations for text or images, such as work \cite{hao2024optimizing}.
\textbf{(b)} A feedback model is trained using a large annotated dataset (typical manual score) and then is employed for RL, like PPO, which is employed in \cite{cao2023beautifulprompt, rosenman2023neuroprompts}. \textbf{(c)} Rewards are generated through a \textbf{fixed external LVLM}. The upper bound of evaluation depends on external models. \textbf{(d)} Rewriter and Reward Model share the same weight, achieving self-improvement by an iterative method to generating answers and self-judgment.}
\label{different reward}
\vspace{-4mm}
\end{figure*}

\subsection{Self-Rewarding Learning}
\vspace{-1mm}
Self-Rewarding Learning uses the same model to perform iterative training to realize self-improvement.
Given a sequence pair $(x, y)$, reinforcement learning fine-tuning demands the definition of the reward function $\textbf{r}(\textbf{x},\textbf{y})$ to quantify the value of the response $\textbf{y}$ to the given input $\textbf{x}$. The objective of the RL fine-tuning can be defined as:
\vspace{-1mm}
\begin{equation}
\begin{aligned}
    \mathcal{L}_{RL}(\theta) &= \mathbb{E}_{\textbf{x}\sim q(\cdot),\textbf{y}\sim p_\theta(\cdot|\textbf{x})}[r(\textbf{x},\textbf{y})]\\
    &-\lambda\mathbb{E}_{\textbf{x}\sim q(\cdot)}{\rm KL}(p_\theta(\cdot|\textbf{x})||p_{ref}(\cdot|\textbf{x}))
    \label{eq4}
\end{aligned}
\end{equation}
where the KL regularization enforces the policy model $p_\theta$ to be close to the reference model $p_{ref}$, and $\lambda$ is set to control the deviation between policy model $p_\theta$ and the reference model $p_{ref}$.

Assuming a pair of responses $<\textbf{y}^1,\textbf{y}^2>$, with a human annotator labeling one of them to be more aligned with human preferences, denoted as $\textbf{y}^w \succ \textbf{y}^l | x$. the Bradley-Terry (BT) \cite{brown2020language} model stipulates that the human preference distribution $p^*$ can be written as follows:
\vspace{-1mm}
\begin{equation}
    \!\!p^*(\textbf{y}^1 \!\succ\! \textbf{y}^2 | \textbf{x}) \!= \!\frac{{\rm exp}(r(\textbf{x},\textbf{y}^1))}{{\rm exp}(r(\textbf{x},\textbf{y}^1))\!+\!{\rm exp}(r(\textbf{x},\textbf{y}^2))}\!
\end{equation}

In self-rewarding training, the same LVLM is used to produce reward $\textbf{r}=[r_1,r_2,...,r_l]$, so  the conditional probability distribution $p_\theta(r|(\textbf{x},\textbf{y}))$ can be expressed as follows:
\vspace{-1mm}
\begin{equation}
    p_\theta(\textbf{r}|(\textbf{x},\textbf{y})) = \prod_{k=1}^l{(r_k|(\textbf{x}, \textbf{y}),\textbf{r}_{<k})}
    \label{eq3}
\end{equation}
where $\textbf{r}_{<1}$ is usually null or a start token and $\textbf{r}_{<k}=[r_1,r_2,..,r_{l-1}]$, $k \in \{2,...,l\}$.

\subsection{Related Work}\label{sec:related_work}
\vspace{-1mm}
\begin{figure*}[!t]
\centering
    \includegraphics[width=1\linewidth]{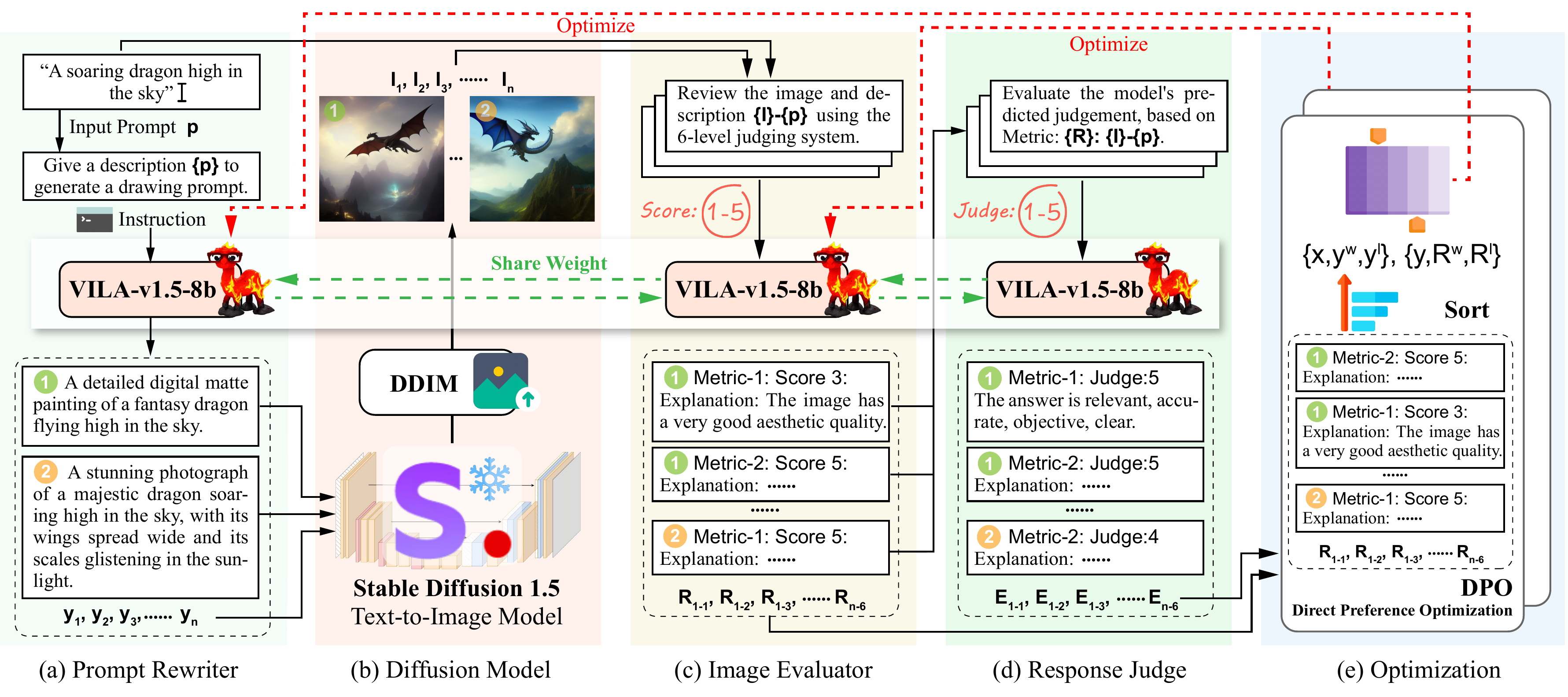}
    \vspace{-7mm}
    \caption{\small \textbf{The overall framework in our prompt optimizing framework.} It involves five steps, arranged from left to right, (\textbf{a) Prompt Rewriter} sample multiple candidates $\textbf{y}$, \textbf{(b) Diffusion Model} generates images from the candidates, \textbf{(c) Image Evaluator} act as image evaluate models, to generate image evaluate responses $\textbf{R}$, \textbf{(d) Response Judge} act as judge models to judge the response from evaluator and get response evaluation $\textbf{E}$, and \textbf{(e) Optimization} with the response from evaluator and judge, and then update the LVLM. }
    \vspace{-3mm}
    \label{self-training}
\end{figure*}

\noindent{\textbf{Prompt Engineering.}}
\citet{hao2024optimizing} propose a prompt adaption framework for prompt engineering. To implement reinforcement learning fine-tuning, a reward function for image aesthetics and alignment is defined. 
Bestprompt~\cite{pavlichenko2023best} is proposed to detect keywords by genetic algorithm and then form prompts to obtain the better aesthetics of images.
Beautiful-Prompt~\cite{cao2023beautifulprompt} first trains two reward models: Aesthetics and PickScore with a large dataset and then optimizes the language model using PPO.
NeuroPrompt~\cite{rosenman2023neuroprompts} utilizes constrained text decoding with a pre-trained language model to produce prompts. \cite{datta2023prompt} is proposed as a Prompt Expansion framework to improve the diversity in text-to-image generation. 
Inspired by visual language modeling, recently, more and more methods \cite{li2022vision, liu2024visual, chen2024mllm,yang2023dawn,yu2023mm} attempt to use a prior knowledge of LVLM to analyze images.
The vision-language model is a large multimodal model, which bridges the gap between language and images. CLIP~\cite{radford2021learning} is trained with large-scale paired text and images using contrastive learning. In this way, the pre-trained LVLMs capture rich vision-language correspondence knowledge. ALIGN \cite{jia2021scaling} scales up the training process, using the larger images-text pairs but noisy data. Recently, with the great success of Large Language Models (LLMs), some work has been devoted to enabling LLMs to use image inputs. OpenFlamingo~\cite{awadalla2023openflamingo} and LLaMA-Adapter \cite{zhang2023llama} construct multimodal models based on the best LLM LlaMA \cite{touvron2023llama}. To further improve the model’s instruction-following abilities, LlaVA~\cite{liu2024visual} employs visual instruction tuning that yields promising results. ViLA~\cite{lin2024vila} achieves multi-image reasoning through a better training strategy.

\noindent{\textbf{Text-to-Image generation.}}
Text-to-image models usually refer to the generative model which synthesizes an image from a given text. Earlier work, GAN \cite{reed2016generative,tao2022df} and VAE \cite{ramesh2021zero, ding2021cogview} have been extensively studied in this field.
Recently, the diffusion-based models \cite{rombach2022high, gu2022vector} further improve the quality of text-to-image generation. Given the generated output is no longer a serious concern, exploring how to optimize the prompt to maximize the potential of generative models becomes the primary object of our research. Several works \cite{hao2024optimizing,cao2023beautifulprompt} have been put into unlocking the maximum potential of text-to-image models by optimizing prompts using large language models. Besides, some reinforcement learning-based methods have gained consistent improvements in changing the output preferences of large language models. Reinforcement Learning from Human Feedback (RLHF) \cite{yuan2024rrhf} yields promising results in various NLP tasks \cite{christiano2017deep, ibarz2018reward,stiennon2020learning,jaques2019way}, at the cost of collecting large amounts of human feedback. 
Proximal Policy Optimization (PPO) \cite{schulman2017proximal} is optimized according to the reward signal rather than human feedback and remains stable by approximating the old and new strategies.
With these challenges in mind, Direct Preference Optimization (DPO) \cite{rafailov2024direct} avoids the drawback of large amounts of human-annotated data to train reward models by using its language model as a reward model. The different strategies to provide rewards are shown in \Cref{different reward}.


\section{Methodology}
\vspace{-2mm}
In this section, we first describe the pipeline of dataset construction (Section~\ref{sec:dataset}), and how to initialize LVLM with both prompt optimization and preference assessing capabilities (Section~\ref{sec:init}) and then detail self-rewarding training algorithms (Section~\ref{sec:self-reward}). Next, we introduce the use of AI feedback for RL in self-rewarding training (Section~\ref{sec:RLAIF}) and how to transform LVLM into a reward model by LVLM-as-a-judge (Section~\ref{sec:VLM-as-a-judge}).

\subsection{Dataset Construction Pipeline}\label{sec:dataset}
\vspace{-1mm}
\noindent{\textbf{Prompt rewrite data.}}
To equip the LVLM with initial rewriting capabilities, rewrite prompts data is first given to train in the supervised fine-tuning (SFT) manner. This type of data is presented in pairs, i.e., \{raw prompt, rewritten prompt\}, to guide the LVLM to rewrite prompt. Prompt rewrite data contains 104,487 prompt pairs in \cite{cao2023beautifulprompt}. 

\noindent{\textbf{Evaluation data.}}
The LVLM specializes in image understanding and analyzing \cite{zhu2024minigpt, li2022vision}. However, image aesthetics is still a subjective understanding and not easy for models.
Although unnecessary, LVLM appears more difficult to score images and construct preference pairs without fine-tuning on evaluation data \cite{yuan2024self}.
Therefore, evaluation data is constructed to offer the model some examples to understand the aesthetics. The prompt data in the previous paragraph is used to generate images by diffusion model \cite{rombach2022high}. These images are then scored using the prompt template in the LVLM-as-a-judge way \cite{zheng2024judging, chen2024mllm}. We use a subset of the aesthetics training set and the Pickapic-v1 training set. The paired images and evaluation prompt are both fed into VILA-40B as well as the facts to obtain the response as truth response for training.
We add a truth cue to the evaluation prompt, if the VILA-40B's scoring of the two images is consistent with the facts, the evaluation response is retained. 

\noindent{\textbf{Judgment data.}}
To improve the LVLM's evaluation capabilities for responses \cite{zheng2024judging,zhu2023judgelm}, we also create a dataset for the model to understand which responses to aesthetics or alignment are more reasonable. 
Specifically, we construct data that assign higher scores to reasonable responses and lower scores to irrelevant responses, which is some kind of confidence.
To improve the evaluation ability of LVLM, we create a new dataset to evaluate the model's responses to aesthetics and PickScore of the images. 
Similarly, we feed the standard (i.e., VILA-40B's response) and irrelevant answers into VILA-40B, and if the model has a higher score in the standard answer than in the irrelevant answer, the answer remains in the dataset. A common observation is that most images generated from rewritten prompts are typically assigned scores of 3 or 4. Therefore, we also include some raw prompts to maintain a more uniform score distribution.

\subsection{Initialization LVLM Capability of Rewriting and Assessing Preference}\label{sec:init}
\vspace{-1mm}
After we obtain the initial dataset, supervised fine-tuning (SFT) is employed to initiate the model with the ability of image evaluation and response judgment. 
Therefore, we consider the input $\textbf{x}$ for a specific task to be derived from the distribution $q(\cdot)$. Meanwhile, the probability distribution in the SFT training data can be represented as $p_{data}(\cdot|\textbf{x})$. The object function of SFT can be represented as:
\begin{equation}
    \mathcal{L}_{SFT}(\theta) = -\mathbb{E}_{\textbf{x}\sim q(\cdot),\textbf{y}\sim p_{data}(\cdot|\textbf{x})}[{\rm log}p_\theta(\textbf{y}|\textbf{x})]
    \label{eq2}
\end{equation}
Since the SFT training data has high-quality labeled responses $\textbf{y}$, the main purpose of this object function is to approximate the model's predictive distribution $p_\theta(\textbf{y}|\textbf{x})$ to the target $p_{data}(\textbf{y}|\textbf{x})$. 

\subsection{Training with Self-Rewarding}\label{sec:self-reward}
\vspace{-1mm}
The main objective of the self-rewarding model is to complete iterative training with limited amounts of human-labeled data on the pre-trained model. During an iteration, the model generates a solution and a judgment on this solution. Unlike other methods that use a fixed reward model \cite{ouyang2022training}, the self-rewarding model uses the model of the current iteration to generate the rewards. 
Therefore, the model can both improve its generative capabilities when acting as a generative model and get its boost when acting as a rewarding model, since the corresponding responses are generated by the same mechanism \cite{yuan2024self}.

The whole training process is concluded as follows: the model first starts with pre-trained LVLMs (denoted as $\VLM_{init}$), training with prompt rewrite data and evaluation data in SFT manner, resulting in $\VLM_{SFT}$.
To further enhance the model's performance, we sample multiple candidates from the raw prompt and construct preference pairs based on the LVLM scores, and then the model is trained in the DPO manner. This training process is iterated M times, yielding models denoted as $\{\VLM_{\DPO_1}, \VLM_{\DPO_2}, \cdots,\VLM_{\DPO_M}\}$.
The bottleneck of prompt engineering can be summarized into two aspects: 1) it is difficult to align with human preference and thus a large number of human-labeled data is required; 2) the whole training process is too complex, typically requiring a larger trained reward model. 
Therefore, we propose a self-rewarding prompt rewriting model to obtain better performance through AI self-feedback and an evolving training process.
The overall framework is shown in Fig.~\ref{self-training}. 
First, the model acts as a solver sampling multiple candidate answers from raw prompts. These candidates are then fed into a fixed text-to-image model to generate the corresponding images. Again, the previous solver is used as the reward model to generate responses, for scoring these images generated from candidates. The details of scoring are discussed below. Finally, preference pairs are constructed for model DPO training.

\subsubsection{Training on Rewriting}\label{training_rewrite}
\vspace{-1mm}
We consider an LVLM to be parameterized by $\theta$ and denoted by $p_\theta$. The model takes a prompt as sequence $\textbf{x} = [x_1, x_2, ..., x_n]$ to generate the corresponding response $\textbf{y} = [y_1, y_2, ..., y_m]$. The response $\textbf{y}$ is thus regarded as a sample from the conditional probability distribution $p_\theta(\cdot|\textbf{x})$. 
Specifically, $x_i$ and $y_j$ represent the tokens from the same predetermined vocabulary within the sequences $\textbf{x}$ and $\textbf{y}$, respectively. To generate the $y_j$ for a given position, the auto-regressive model $p_\theta$ exploits the previously generated tokens to generate subsequent tokens up to the maximum length or the end token. Therefore, the conditional probability distribution $p_\theta(\textbf{y}|\textbf{x})$ can be expressed as:
\begin{equation}
    p_\theta(\textbf{y}|\textbf{x}) = \prod_{j=1}^m{(y_j|\textbf{x},\textbf{y}_{<j})}
    \label{eq1}
\end{equation}
where $\textbf{y}_{<1}$ is usually null or a start token and $\textbf{y}_{<j}=[y_1,y_2,..,y_{j-1}]$, $j \in \{2,...,m\}$.

\subsubsection{Training on Preference Assessing}
\vspace{-1mm}
Beyond the model training on prompt rewriting, we also perform preference assessing training on the same model to improve its assessing ability. Similar to Sec.~\ref{training_rewrite}, the model generates the judgment on the response $\textbf{R}$ for a response $\textbf{y}$ on prompt and the generated image. The conditional probability distribution can be defined as:
\begin{equation}
    p_\theta(\textbf{R}|\textbf{y}) = \prod_{k=1}^n{(R_j|\textbf{y},\textbf{R}_{<k})}
    \label{eq1_}
\end{equation}
where $\textbf{y}_{<1}$ is usually null or a start token and $\textbf{R}_{<k}=[R_1,R_2,..,R_{k-1}]$, $j \in \{2,...,n\}$.

Therefore, RL fine-tuning in self-rewarding can be optimized with the loss function, i.e.,
\begin{align}
\mathcal{L}_{RL}(\theta_{t+1}) &= \mathbb{E}_{\textbf{x}\sim q(\cdot),\textbf{y}\sim p_{\theta_t},\textbf{r}\sim p_{\theta_t}(\cdot|\textbf{(x,y)})}[r(\textbf{x},\textbf{y})]\notag \\ 
&-\lambda\mathbb{E}_{\textbf{x}\sim q(\cdot)}{\rm KL}(p_\theta(\cdot|\textbf{x})||p_{ref}(\cdot|\textbf{x})) \notag\\
& + \mathbb{E}_{\textbf{y}\sim q(\cdot),\textbf{R}\sim p_{\theta_t},\textbf{E}\sim p_{\theta_t}(\cdot|\textbf{(y,R)})}[E(\textbf{y},\textbf{R})] \notag\\ 
&-\lambda\mathbb{E}_{\textbf{y}\sim q(\cdot)}{\rm KL}(p_\theta(\cdot|\textbf{y})||p_{ref}
\label{eq5}
\end{align}
where $\theta_t$ denotes the $t$-th parameters of the model, the $r(\textbf{x},\textbf{y})$ and the $E(\textbf{y},\textbf{R})$ denote the reward function of the prompt $x$ and the response $y$, respectively.
Since the self-reward model is an iterative model, in which the iterative process results in a series of models with different weights of the same structure. 
For better understanding, the parameter of the result model by \Cref{eq2} is denoted as $\theta_0$, while the parameters of the result models optimized through \Cref{eq5} are denoted as $\{\theta_1, \theta_2, \cdots\}$.
As the model's parameters continue to be optimized for the ability to follow instructions, so does the reward ability, thereby facilitating self-improvement.

\subsection{RL from AI Feedback}\label{sec:RLAIF}
\vspace{-1mm}
\noindent{\textbf{AI Feedback.}}
One of the main investigations of our work is how to use prior knowledge of large models to guide text-to-image models.
Unlike time-consuming and labour-intensive human feedback training, AI feedback \cite{lee2023rlaif} training can provide reward signals for a given task through its own knowledge. This approach usually requires two models, one acting as a solver of the downstream task, and one acting as a judge of the solver. It is feasible to apply an external language model as a judge (or reward model) or to use only itself  (i.e. a model that acts as both a solver and a judge) to achieve self-improvement in a specific task. 

Therefore, preference pairs $\langle$ raw prompt $x$, winner prompt $y^w$, loser prompt $y^l$ $\rangle$ need to be constructed to train the model. 
The model first generates different $N$ candidates from randomly selected raw prompts (in the previous data). After generating images using a fixed text-to-image model, the images and evaluation prompts are input into the model for scoring. Next, the prompts corresponding to the highest and lowest-scored images are treated as winners and losers, respectively. In addition, the image evaluation responses are judged by LVLM and the pairwise data for DPO are also constructed.
The model is then tuned with DPO \cite{rafailov2024direct}.

\noindent{\textbf{DPO.}} Assuming access to a static dataset of comparisons $\mathcal{D}=\{\textbf{x}_i, \textbf{y}^w_i, \textbf{y}^l_i\}$, which is sampled from $p^*$. The optimal RLHF policy $\pi^*$ under the Bradley-Terry model satisfies the preference model:
\begin{align}
    p^*(\textbf{y}^1 \succ \textbf{y}^2 | \textbf{x}) = \Big[1 &+ \exp\left(\lambda\log\frac{\pi^* (\textbf{y}^2|\textbf{x})}{\pi_{ref}(\textbf{y}^2|\textbf{x})}\right) \notag\\
    &- \lambda\log\frac{\pi^*(\textbf{y}^1|\textbf{x})}{\pi_{ref}(\textbf{y}^1|\textbf{x})})\Big] ^{-1}
\end{align}
Given a preference pair $\langle\textbf{x}$, $\textbf{y}^w$, $\textbf{y}^l\rangle$, the object function of DPO is to seek a maximum likelihood of the parameterized policy $\pi_\theta$ by reference model $\pi_{ref}$. 
\begin{align}
&\mathcal{L}_{DPO}(\pi_\theta;\pi_{ref}) \notag\\
&= -\mathbb{E}_{(\textbf{x}, \textbf{y}^w, \textbf{y}^l)\sim\mathcal{D}} 
\Big[
    {\rm log}\sigma\big(
        \Delta_\lambda(\textbf{y}^w|\textbf{x}) 
        - \Delta_\lambda(\textbf{y}^l|\textbf{x})
    \big)
\Big], \notag \\
&\text{where}~~\Delta_\lambda(y|x) = \lambda \log \frac{\pi_\theta(y|x)}{\pi_{ref}(y|x)}. \label{eq:dpo_loss}
\end{align}
The $\lambda$ is a parameter controlling the deviation from the base reference policy $\pi_{ref}$.

\subsection{LVLM-as-a-judge for Aesthetics and Rewarding} \label{sec:VLM-as-a-judge}
\vspace{-1mm}
To improve the text-to-image models, we consider this from two perspectives: \textbf{the aesthetics of the generated image} and \textbf{the capability to follow instructions}, respectively. Hence, to empower the model in generating appropriate reward signals for the images, we establish a judging template comprising aesthetic score, pick score and alignment score. This enables the LVLM to evaluate the generated images effectively. Details of prompts for LVLM-as-a-judge can be found in Appendix.

\noindent{\textbf{Aesthetic Score.}}
Aesthetics are assessed by prompting the model to consider aspects such as composition, color, lighting, and visual appeal. To facilitate judgment-making, we design a grading system that allows the model to attribute a certain grade to the generated images, thus obtaining the corresponding score. For each evaluation, the model is asked to provide both the score and a brief explanation, employing a chain-of-thought reasoning approach.


\noindent{\textbf{PickScore \cite{kirstain2024pick}.}}
PickScore is an important metric to measure human preferences. We consider whether the image represents the given text in a way that is favored by humans. Nonetheless, it is difficult for the model to understand human preferences directly, so we included evaluation data (details in Section~\ref{sec:init}) for model initialization.

\noindent{\textbf{Relevance Score.}}
The judgment template for the relevance score uses the same hierarchical form as the aesthetic score.
The key difference is that it evaluates the model's attention to the user instructions for text-to-image generation in three areas: presence of the object, accurate count of the object, and correct relationships between objects. 
\vspace{-1mm}
\section{Experiments}
\vspace{-1mm}

\subsection{Settings}
Experiments are conducted on the public text-to-image model Stable Diffusion v1.5, and the denoising steps are set to 20 to accelerate the image sampling.
The base model we employ in this paper is the smaller model VILA-v1.5-8b~\cite{lin2024vila}.  
For SFT, we use AdamW optimizer ($\beta_1=0.9, \beta_2=0.95$), with a batch size of 16 and a weight decay of 0.1. We use an initial learning rate of 2e-5, with a linear warm-up and cosine decay schedule. 
To improve the diversity of the candidates, we sample $N=8$ candidates with temperature $T=0.9,p=0.9$ from one raw prompt. When validating these candidates with the reward model, we utilize a temperature $T=0.9,p=0.9$, sampling three times and averaging the scores to determine the final score for further DPO training. The overall score is calculated by summing up the aesthetics score, pick Score and the relevance score. The highest one and the lowest one are kept as winners and losers but discarded if they have the same score. We perform two DPO iterations and the size of each prompt rewrite training data is 10k and 20k, respectively. Besides, 10k judgment data is employed to improve image evaluation.
In the stage of DPO, we employ AdamW with an initial learning rate of {1e-5, 5e-6} without weight decay. The batch size is set to 32 and $\beta$ in DPO is set to 0.1. 
The model is trained for 4 epochs at each iteration. All experiments are conducted on $4 \times$ NVIDIA A800 80G GPUs.
For evaluation, we adopt beam search \cite{vijayakumar2016diverse} with a beam size of 4 and a length penalty of 1.0.

\begin{table}[t]\small
\centering
\resizebox{\linewidth}{!}{
\setlength{\tabcolsep}{3.2pt}
\begin{tabular}{lcccc}
    \toprule
    \textbf{Method} & \textbf{$\sD_{rl}$} & \textbf{PickScore} & \textbf{Aes.} & \textbf{CLIP} \\
    \midrule
    Original & - & 20.74 & 5.50 & \textbf{0.27} \\
    \midrule
    MagicPrompt~\cite{Magicprompt} & - & 20.11 & 5.79 & 0.22 \\
    ChatGPT~\cite{chatgpt} & -& 20.73 & 5.92 & 0.25 \\
    Beautiful-Prompt~\cite{cao2023beautifulprompt}  & 40k & 20.84 & 6.52 & 0.24 \\\hline\midrule
\rowcolor{gray!15}\multicolumn{5}{c}{\textit{Our Method}}  \\\midrule
    $\VLM_{\SFT}$ & - & 20.79 & 5.95 & 0.25 \\
    $\VLM_{\DPO_1}$ & 10k & 20.81 & 6.31 & 0.24 \\
    $\VLM_{\DPO_2}$  & 20k & \textbf{20.86} & \textbf{6.59} & 0.24 \\
     \bottomrule
\end{tabular}}
\vspace{-2mm}
\caption{\small Evaluation of the aesthetic score and CLIP score on Beautiful-Prompt test set. ``$\sD_{rl}$'' means the size of the train set we used in reinforcement learning. ``Aes.'' and ``CLIP'' mean the aesthetic and CLIP scores, respectively.}
\label{main_experience_on_bp}
\vspace{-4mm}
\end{table}

\begin{table}[t]\small
\centering
\resizebox{\linewidth}{!}{
\setlength{\tabcolsep}{4pt}
\begin{tabular}{lcccc}
\toprule
\textbf{Method} & \textbf{Type} & \textbf{Aes.} & \textbf{CLIP} \\
\midrule
Original$\dag$ & Human & 5.47 & \textbf{0.28} \\
Human Engineered Prompt$\dag$ & Human & 5.87 & 0.26 \\
\midrule
NeuroPrompts\cite{rosenman2023neuroprompts} & AI & 6.27 & - \\
Promptist\cite{hao2024optimizing} & AI & 6.26 & 0.26 \\
\rowcolor{gray!15}$\VLM_{\DPO_2}$ (Ours)  & AI &  \textbf{6.57} & 0.26 \\\bottomrule        
\end{tabular}}
\vspace{-2mm}
\caption{\small Evaluation of the aesthetic score and CLIP score on DiffusionDB. Values marked with $\dag$ from \cite{hao2024optimizing}. }
\label{main_experience_on_diffusionDB}
\vspace{-2mm}
\end{table}

\vspace{-1mm}
\subsection{Comparative Methods}
We compare our method with the following approaches: MagicPrompt~\cite{Magicprompt}, ChatGPT~\cite{chatgpt}, Beautiful-Prompt~\cite{cao2023beautifulprompt}, NeuroPrompts~\cite{rosenman2023neuroprompts} and Promptist~\cite{hao2024optimizing}. It is worth noting that, we prompt ChatGPT~\cite{chatgpt} to generate an expansion of the user-provided prompt for generative models. Other models generate results using their open-source weights. The ``Human Engineered Prompt'' refers to the prompts written by humans, while its simplified version of these prompts is used as the original prompt. In real-world usage, users only need to provide simple prompt to obtain high-quality results.

\subsection{Results}
The model is validated on two datasets, Beautiful-Prompt \cite{cao2023beautifulprompt} and DiffusionDB \cite{wang2023diffusiondb} (100k), respectively. As shown in Table~\ref{main_experience_on_bp}, our method outperforms other methods. This performance gap is more evident on larger datasets DiffusionDB (100k). As can be seen in Table~\ref{main_experience_on_diffusionDB}, our method achieves a performance gain compared to other methods. Compared to the most competitive method Beautiful-Prompt, our method achieves a 0.1 improvement in the Aesthetic Score, while maintaining no degradation in the PickScore and CLIP score. However, on the larger DiffusionDB test set, our method achieves a 0.3 improvement compared to the Neuroprompt \cite{rosenman2023neuroprompts} and Promptist \cite{hao2024optimizing}.

\begin{table}[t]\small
\centering
\vspace{-1mm}
\setlength{\tabcolsep}{4pt}
\begin{tabular}{ccc|c|ccc}
    \toprule
    \textbf{Pick} & \textbf{Aes.}  & \textbf{Align} &\textbf{Reward} & \textbf{PickScore} & \textbf{Aes.}  & \textbf{CLIP}  \\
    \midrule
\checkmark & \checkmark & \checkmark& Self & \textbf{20.86} & 6.59 & \textbf{0.24} \\
    & \checkmark  & \checkmark& Self & 20.73 & 6.59 & \textbf{0.24} \\
    \checkmark & \checkmark & & Self & 20.79 & \textbf{6.61} & 0.22 \\
    \checkmark &  & \checkmark& Self & 20.76 & 6.49 & \textbf{0.24} \\
\checkmark & \checkmark & \checkmark& Fixed & 20.73&6.26&\textbf{0.24}\\
     \bottomrule
\end{tabular}
\vspace{-2mm}
\caption{\small Ablation study on different prompts and reward mode. Checkmark is enabled prompts in image evaluation.}
\vspace{-4mm}
\label{ablation}
\end{table}

\subsection{Ablation Study}
We perform ablation experiments from two perspectives: how the prompt affects the LVLM as a reward model and how self-rewarding training improves the model.
First, as can be seen in Table~\ref{ablation}, the model without Pick Prompt for rewarding received a high aesthetic score but leads a significant drop in PickScore. It also suggests that some shifts exist between human preferences and image aesthetics. In addition, it can be observed a reduction of the aesthetic score and CLIP score when missing the aesthetic prompt or relevance prompt. 
Second, we compare the performance between the model using self-rewarding and fixed-rewarding. Self-rewarding outperforms the fixed-rewarding methods. More discussion about the self-rewarding method can be found in the Appendix~\ref{app:prompt}.

\begin{figure}[!t]
    \centering
    \includegraphics[width=0.9\linewidth]{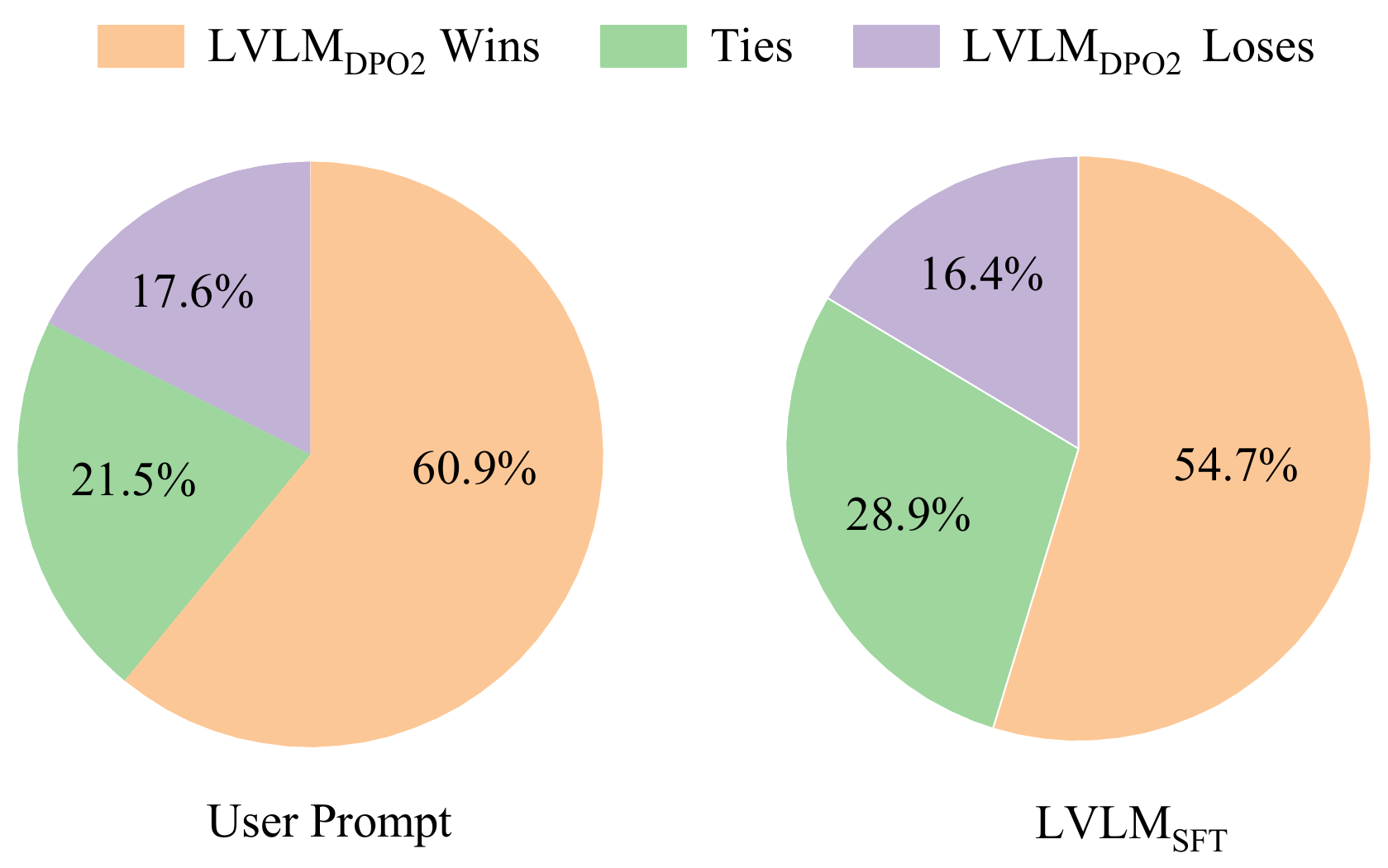}
    \vspace{-2mm}
    \caption{\small Human evaluation results. The result of $\VLM_{DPO_2}$ are more preferred by human compared with the result of User Prompt and $\VLM_{SFT}$.}
    \label{fig:human}
    \vspace{-5mm}
\end{figure}

\begin{figure}[!t]
    \centering
    \includegraphics[width=\linewidth]{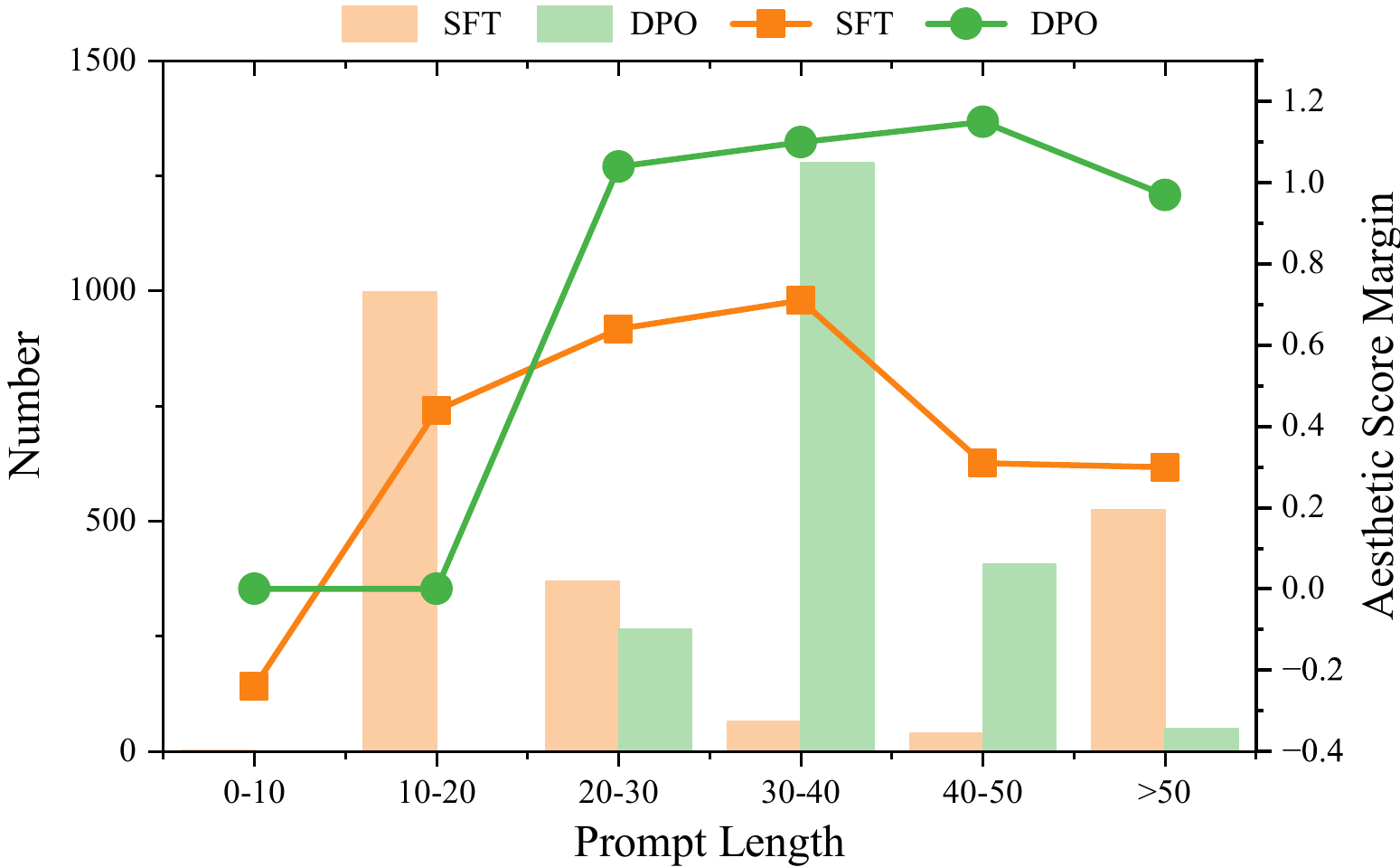}
    \vspace{-6mm}
    \caption{\small Generated prompt length (bar graph) and aesthetic score improvement (line graph) compared with raw prompt.}
    \label{fig:len}
    \vspace{-3mm}
\end{figure}

\subsection{Human Evaluation}
To better compare the existing methods with our model from the human perspective, a human study is conducted. We randomly select prompts and obtain the optimized prompt through different models, and then the images are generated. Then, 20 volunteers are asked to rank the different images. Each volunteer is presented with two images at a time and asked to select the one they find more appealing.
The volunteers in this study are randomly selected from a pool of individuals with diverse educational backgrounds. 
The result is shown in Figure~\ref{fig:human}.
It can be observed that the images obtained through the optimized prompt generated by our model are selected most frequently by the volunteers. And the DPO model is better than the SFT baseline, which achieves about 55\% win rate.

\begin{figure*}[!t]
    \centering
    \vspace{-2mm}
    \includegraphics[width=1\linewidth]{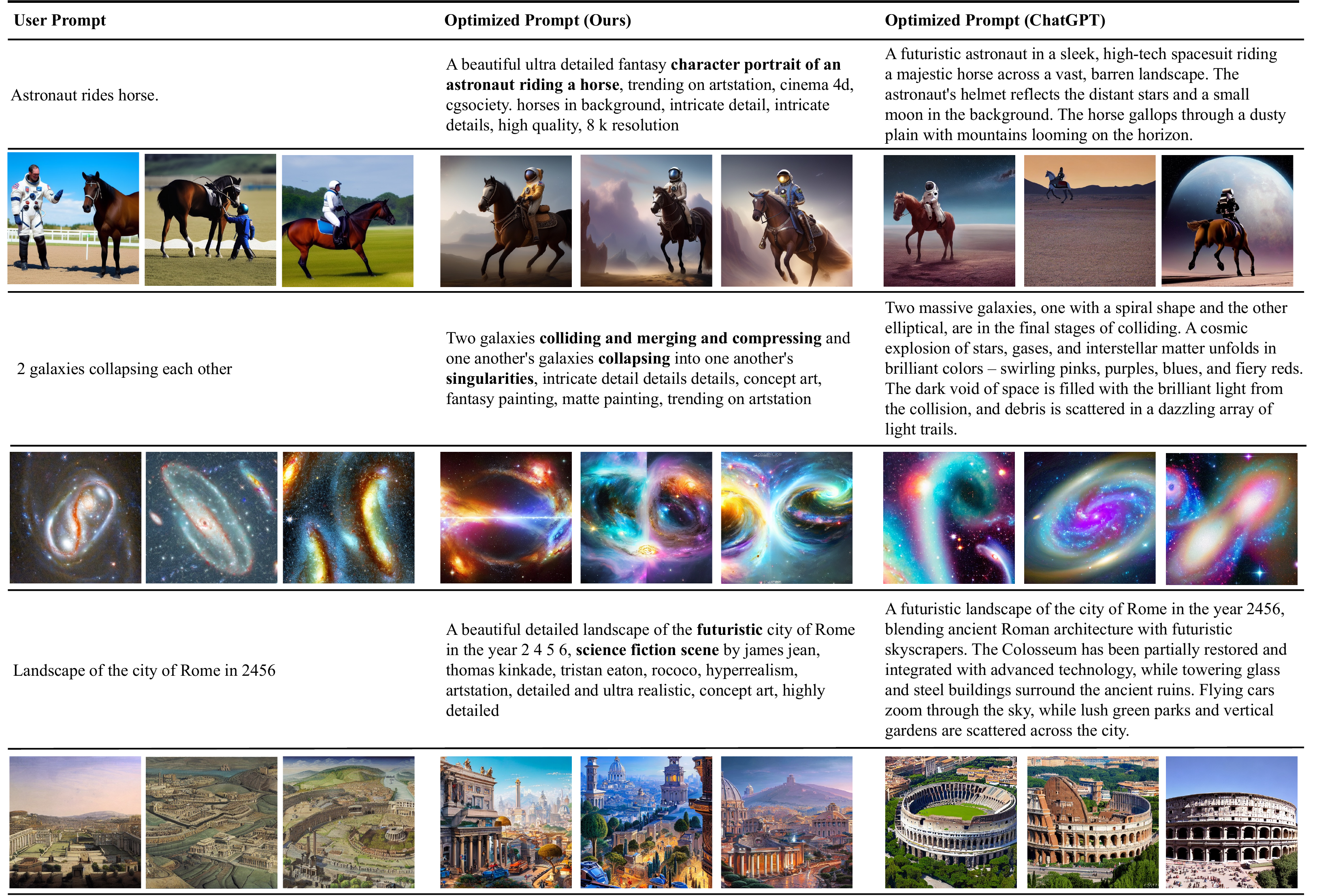}
    \vspace{-7mm}
    \caption{\small The generated images with the optimized prompts using our method}
    \label{fig:visual_result}
    \vspace{-2mm}
\end{figure*}

\begin{table}[t]\small
\centering
\setlength{\tabcolsep}{5pt}
\begin{tabular}{cccccc}
\toprule
\multicolumn{6}{c}{\textbf{\# Training Data}} \\
\midrule
\multicolumn{2}{c}{Stage} & \multicolumn{2}{c}{Beautiful-Prompt} & Promptist & Ours \\\midrule
\rowcolor{gray!15} \multicolumn{2}{c}{SFT} & \multicolumn{2}{c}{143k} & 360k & 156k \\
\multicolumn{2}{c}{{\textit{Rewrite}}} & \multicolumn{2}{c}{143k} &  360k & 104k \\
\multicolumn{2}{c}{\textit{Evaluation}} & \multicolumn{2}{c}{-} &  - & 32k \\

\multicolumn{2}{c}{\textit{Judgment}} & \multicolumn{2}{c}{-} &  - & 20k \\
\midrule
\rowcolor{gray!15}\multicolumn{2}{c}{RM} & \multicolumn{2}{c}{7M} & - & - \\
\midrule
\rowcolor{gray!15}\multicolumn{2}{c}{RL} & \multicolumn{2}{c}{40k} & 90k & 40k \\
\multicolumn{2}{c}{\textit{Rewrite}} & \multicolumn{2}{c}{40k} &  90k & \{10k, 20k\} \\
\multicolumn{2}{c}{\textit{Evaluation}} & \multicolumn{2}{c}{-} &  - & 10k \\
\bottomrule
\end{tabular}
\vspace{-2mm}
\caption{\small Training set number on different stages of Beautiful-Prompt\cite{cao2023beautifulprompt}, Promptist\cite{hao2024optimizing}, and ours. \textbf{RM} means the training of the reward model. Our method requires addition data on image evaluation and response judgment for rewarding.}
\label{tab:trainingdata}
\vspace{-3mm}
\end{table}

\begin{table}[!t]\small
\centering
\setlength{\tabcolsep}{3pt}
\begin{tabular}{cccc}
\toprule
Method & $\VLM_{init}$ & $\VLM_{\SFT}$ & $\VLM_{\DPO_1}$ \\
\midrule
Test-Set Accuracy & 48.0\% & 66.1\% &  66.4\% \\
 \bottomrule        
\end{tabular}
\vspace{-2mm}
\caption{\small Evaluation on Pickapic-v1 test set. Our self-rewarding model realizes the ability to self-improvement}
\label{dpo_experience_on_pickapic}
\vspace{-5mm}
\end{table}


\subsection{Further Discussion}
To further understand how the prompt optimization affects the original prompt, we show some analysis in Figure~\ref{fig:len}. Considering the length of prompts has a significant relationship with the quality of generation, we count the length of optimized prompts of $\VLM_{SFT}$ and $\VLM_{DPO_2}$ and their aesthetic score margins compared to the original prompt. An immediate observation is that the length of most optimized prompts is between 30 and 40. Besides, aesthetic scores gradually increased with the length. 
Besides, we also present the size of the training set in Table~\ref{tab:trainingdata}. In SFT stage, we use prompt rewrite data (104k), evaluation data (32k) and judgment data (20k), while in RL fine-tuning stage, we use prompt data {10k, 20k} and new evaluation data {10k} sampled from LVLM. Note that we just expanded the dataset after each iteration, whereas 10k prompt data is the subset of 20k. This means only 20k training data are needed to train this model, while Beautiful-Prompt and Promptist employ the larger training set, i.e., 40k and 90k, respectively. 
To better exemplify the ability of our model to evaluate by self-rewarding, we show the accuracy of the test set in Pickapic-v1 after SFT and DPO training. As shown in Table \ref{dpo_experience_on_pickapic}, our model realizes an increase in evaluation capacity.

\subsection{Quantitative Results}
\vspace{-1mm}
We present more visual results in Figure~\ref{fig:visual_result}, it can be observed that our method rewrites the main content appropriately, and describes the detail, e.g., art style, light, etc. The images generated by these optimized prompts have better aesthetics than the images generated from the original prompts, which are bland visually. In addition, some incorrect prompts can also be corrected. For example, the prompt \textit{``Astronaut rides horse"} mislead the text-to-image models to generate an astronaut standing beside a horse. However, our approach rewrites it as \textit{``...an astronaut riding a horse..."} to make the image more in line with the user's intention. In addition, the model expands some descriptions on some objects, like \textit{``futuristic""} relative to the \textit{``in 2456''}. More results are in the Appendix~\ref{app:quantitative}.

\vspace{-1mm}
\section{Conclusion}
\vspace{-1mm}
In this work, we propose a novel method to optimize prompts for text-to-image models, which can be used to fill the gap between laymen and experts when using a generative model. We first transform the LVLM into a reward model so that it can judge the aesthetics and alignment of the images. In so doing, we can perform AI feedback rather than human feedback. Then, to gain the performance boost based on limited data, we employ self-rewarding training for LVLM. Our model achieves self-improvement through an iterative training approach. Experiments on two datasets show that our method outperforms the other methods.

\section*{Limitations}
The primary limitation of our work is the inability to utilize larger parameter models due to the significant computational resources required. Training and fine-tuning models with billions of parameters, such as VILA-40B, demand substantial GPU memory and processing power, which are often constrained by available hardware. 
Deploying large-parameter models in real-world applications presents considerable challenges.

{
    \small
    \bibliography{main}
}

\clearpage
\appendix 

\section{Comparisons with the common tags}\label{app:tags}
To figure out what word improves the prompt, we count the most frequent words. Evidently, \textit{``fantasy''} and \textit{``intricate''} are popular in the model, which even appears more than once in the same sentence. 

To demonstrate that the performance improvement of the prompt rewriting method is not the result of adding some fixed words, we select the most frequent words and randomly combine them into six tags. Then, we combine it with the original prompt and calculated the scores. The tags are shown in Tab.~\ref{tab:tags} and the results are present in Tab.~\ref{tab:result_tags}.

\begin{table}[h]\small
    \centering
        \begin{tabular}{cr}
            \toprule
            \textbf{Content} & \textbf{Frequency} \\
            \midrule
            fantasy & 2,291 \\
            intricate & 2,146 \\
            portrait & 1,604 \\
            beautiful & 1,211 \\
            highly detailed & 701 \\
            realistic & 596 \\
            high quality & 389 \\
            elegant & 381 \\
            illustration & 121 \\
            \bottomrule
        \end{tabular}
        \caption{\small Analysis of the most frequent words in the optimized prompts in the beautiful-prompt test set.}
        \label{tab:freq}

\end{table}

\begin{table}[h]\small
\centering
\begin{tabular}{cc}
\toprule
\textbf{Tag} & \textbf{Content} \\
\midrule
1 & artstation, highly detailed, elegant \\
2 & 8k, trending on artstation, concept art \\
3 & digital painting, intricate, fantasy \\
4 & illustration, smooth, fantasy \\
5 & portrait, beautiful, illustration \\
6 & realistic, dramatic, high quality \\
\bottomrule
\end{tabular}
\vspace{-2mm}
\caption{\small Combinations of common tags}
\label{tab:tags}
\end{table}

\begin{table}[h]\footnotesize
\centering
\resizebox{\linewidth}{!}{
\setlength{\tabcolsep}{4pt}
\begin{tabular}{crrrrrrr}
\toprule
\textbf{Metric} & \textbf{Tag1} & \textbf{Tag2} & \textbf{Tag3} & \textbf{Tag4} & \textbf{Tag5} & \textbf{Tag6} & \textbf{Ours} \\
\midrule \midrule
PickScore & 20.81 & 20.73 & 20.68 & 20.59 & 20.69 & 20.82 & 20.86\\
Aes. Score & 5.80 & 5.75 & 5.92 & 5.67 & 5.75 & 5.59 & 6.59 \\
\bottomrule
\end{tabular}}
\vspace{-2mm}
\caption{\small The result using different groups of common tags.}
\label{tab:result_tags}
\end{table}

\section{Further Discussion}\label{app:discussion}
An essential mechanism in self-rewarding model is sampling multiple outputs and then judging them to construct preference pairs for RL fine-tuning. Thus, the diversity of candidates greatly affects the performance of self-rewarding, which can be demonstrated from Tab.~\ref{tab:candidates}. It is essential to enlarge the margin between winners and losers with more candidates, which facilitates RL fine-tuning.

\begin{table}\footnotesize
\centering
\resizebox{\linewidth}{!}{
\setlength{\tabcolsep}{1.8pt}
\begin{tabular}{c|ccccc}
\toprule
\textbf{Method} & \textbf{Reward} & \textbf{\#Candidates} & \textbf{PickScore} & \textbf{Aesthetic} & \textbf{CLIP} \\
\midrule
\midrule 
$\VLM_{\SFT}$ & - & - & 20.79 & 5.95 & \textbf{0.25} \\
$\VLM_{\DPO_1}$ & $\VLM_{\SFT}$ & 2 & 20.65 & 6.03 & 0.23 \\
$\VLM_{\DPO_1}$ & $\VLM_{\SFT}$ & 4 & 20.78 & 6.16 & 0.23 \\
$\VLM_{\DPO_1}$ & $\VLM_{\SFT}$ & 8 & 20.81 & 6.31 & 0.24 \\
 \bottomrule
\end{tabular}}
\vspace{-2mm}
\caption{\small Number of candidates}
\label{tab:candidates}
\end{table}

In addition, the self-reward model is at risk of bias or overfitting.  There are many factors that may contribute to model collapse, including data quality and scale, the inherent capabilities of the language model, biases in the generative model, and the number of candidate samples.
To demonstrate the effectiveness of our iterative training, we show the results of one more round of training. We used an additional 5,000 data for one more training. As shown in Tab.~\ref{tab:more_itea}, the third iteration does not cause a significant performance drop.

\begin{table}\small
\centering
\begin{tabular}{c|ccc}
\toprule
\textbf{Method} & \textbf{PickScore} & \textbf{Aes.} & \textbf{CLIP} \\
\midrule
\midrule 
$\VLM_{\SFT}$ & 20.79 & 5.95 & \textbf{0.25} \\
$\VLM_{\DPO_1}$ & 20.81 & 6.31 & 0.24 \\
$\VLM_{\DPO_2}$ & \textbf{20.86} & 6.59 & 0.24 \\
$\VLM_{\DPO_3}$ & \textbf{20.86} & \textbf{6.60} & 0.24 \\
 \bottomrule
\end{tabular}
\vspace{-2mm}
\caption{\small More iteration training.}
\label{tab:more_itea}
\end{table}

\section{More Quantitative Results}\label{app:quantitative}
\begin{figure*}[t]
    \centering
    \includegraphics[width=0.88\linewidth]{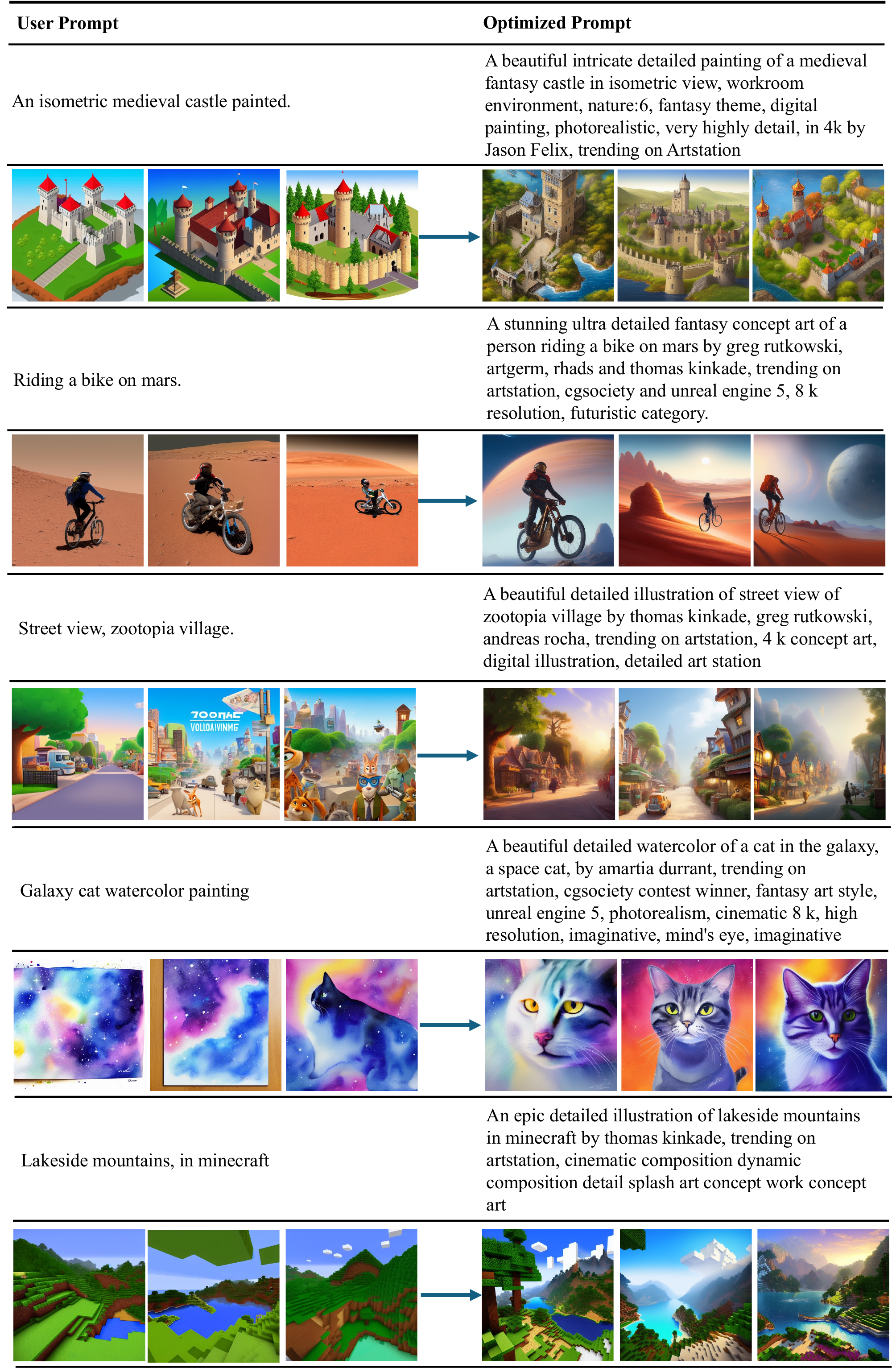}
    \caption{The images generated using our optimized prompt.}
    \label{fig:morevisres}
\end{figure*}

We present more visual results between the images generated with different prompts. As shown in Fig.~\ref{fig:morevisres}, the optimized prompts result in more pleasing images. A more intuitive observation is that the flat, uninteresting view in Minecraft and the more aesthetically pleasing, more detailed view are represented by the optimized prompt before and after optimization, respectively. In addition, the modified prompt has stronger alignment capabilities. For example, the prompt ``\textit{Riding a bike on mars}'' is amended to ``\textit{... a person riding a bike on mars ...}'', and the prompt ``\textit{Galaxy cat}'' is added with the description ``\textit{... a space cat ...}''. 

\section{Prompts for LVLM-as-a-judge}\label{app:prompt}
We present the prompt to transform the LVLM into the reward model. As mentioned in Sec.~\ref{sec:VLM-as-a-judge}, we employ the rating system for image aesthetics, human preferences, and alignment. Note that we input the generated image by original prompt and the image by candidate prompt here to facilitate scoring of the model on the same benchmarks.

For responses judgment, we input two responses to the model that are about how the model evaluated the image in the previous step.



\onecolumn
\begin{tcolorbox}[
        title={Aesthetic Score Prompt},
        halign=left,
        valign=center,
        nobeforeafter,
    ]
Please evaluate the aesthetics of two images (``Image 1'' and ``Image 2'') using the 6-level judging
system described below.

The two images given are independent, and should be evaluated separately and step by step,
ensuring that the order in which the images were presented does not affect your judgment.

\vspace{\baselineskip}
- Poor (Score: 0): The image lacks balance, composition, and visual appeal. Colors may be overly
saturated or dull, causing discomfort to the viewer. Composition is chaotic, distracting, or poorly
framed.

\vspace{\baselineskip}
- Below Average (Score: 1): The image present minimal aesthetic appeal, even if there are
inconsistencies or major flaws in composition, color, lighting, or other aesthetic elements, or make
people feel disjointed or unbalanced, lacking a cohesive visual narrative.

\vspace{\baselineskip}
- Average (Score: 2): The image exhibits adequate aesthetic quality contributes to the image's visual
appeal to some extent but there is room for improvement in terms of creativity or originality or some
aspects of the image may feel generic or uninspired.

\vspace{\baselineskip}
- Above Average (Score: 3): The image has strong aesthetic quality regardless of whether there are
minor imperfections in composition, color, lighting, or other aesthetic elements may still be present
but do not significantly detract from the overall aesthetic, or aesthetic choices may be subjective,
with some viewers preferring different styles or approaches.

\vspace{\baselineskip}
- Very Good (Score: 4): The image is of exceptional aesthetic quality and demonstrates creativity, skill,
and mastery of visual elements even if there is slight room for improvement in composition, color,
lighting, or other aesthetic elements.

\vspace{\baselineskip}
- Excellent (Score: 5): The image is of perfect balance, harmony, and creativity in aesthetics,
creating a visually compelling and impactful image.

\vspace{\baselineskip}
\vspace{\baselineskip}
Please provide a comprehensive explanation of your score.

Note that the score has nothing to do with image input order.

\vspace{\baselineskip}
Output format:

\vspace{\baselineskip}
Output for Image 1:

Score: $<$Your Score for Image 1$>$

Explanation: $<$detailed judgment of Score for Image 1$>$

\vspace{\baselineskip}
Output for Image 2:

Score: $<$Your Score for Image 2$>$

Explanation: $<$detailed judgment of Score for Image 2$>$
\end{tcolorbox}

\begin{tcolorbox}[
        title={Alignment Score Prompt},
        halign=left,
        valign=center,
        nobeforeafter,
    ]
Please evaluate the alignment of two pictures (``Image 1'' and ``Image 2'') to the text (``Text'') using the 6-
level judging system described below.

The two images given are independent, and should be evaluated separately and step by step, ensuring that the
order in which the images were presented does not affect your judgment.

You need to first consider what appears in the image, then whether what is described in the text appears in the
image, and finally give a score based on the system.

\vspace{\baselineskip}
Judging system:
\vspace{\baselineskip}

- No Match (Score: 0): The image does not contain any of the objects or elements mentioned in the text. There is no recognizable connection between the text and the image.

\vspace{\baselineskip}
- Poor Match (Score: 1): The image contains one or a few of the objects mentioned in the text, but these are
peripheral and do not capture the primary content or relationships described. Quantitative relationships are
ignored or inaccurately represented.

\vspace{\baselineskip}
- Fair Match (Score: 2): Some of the primary objects mentioned in the text are present in the image, and at least
one quantitative relationship or object relationship is correctly depicted. However, several key objects or
relationships are missing or inaccurately represented.

\vspace{\baselineskip}
- Good Match (Score: 3): The majority of the objects mentioned in the text are present, and many of the
described quantitative relationships and object relationships are accurately depicted. Minor details may be
missing or slightly altered.

\vspace{\baselineskip}
- Excellent Match (Score: 4): Nearly all objects described in the text are accurately represented in the image,
including precise quantitative relationships and interactions between objects. Only trivial discrepancies or
omissions are present, which do not significantly impact the overall accuracy.

\vspace{\baselineskip}
- Perfect Match (Score: 5): The image perfectly matches the text in terms of the presence of all described
objects, accurate quantitative relationships, and the exact relationships between objects. Every detail mentioned
in the text is present and correctly depicted in the image.

\vspace{\baselineskip}
Text: \textcolor{red}{$<$PROMPT$>$}

\vspace{\baselineskip}
Please provide a comprehensive explanation of your score.
Note that the score has nothing to do with image input order.

\vspace{\baselineskip}
Output format:

\vspace{\baselineskip}
Output for Image 1:

Score: $<$Your Score for Image 1$>$

Explanation: $<$detailed judgment of Score for Image 1$>$

\vspace{\baselineskip}
Output for Image 2:

Score: $<$Your Score for Image 2$>$

Explanation: $<$detailed judgment of Score for Image 2$>$
\end{tcolorbox}

\begin{tcolorbox}[
        title={Pick Score Prompt},
        halign=left,
        valign=center,
        nobeforeafter,
    ]
Please evaluate how well these two images (``Image 1'' and ``Image 2'') generated based on the text (``Text'') are
preferred by humans using the 6-level judging system described below.

The two images given are independent, and should be evaluated separately and step by step, ensuring that the
order in which the images were presented does not affect your judgment.

In this system, 'attractiveness' refers to the visual appeal of an image to the human in terms of color, composition,
lighting, style, and detail.

\vspace{\baselineskip}
Judging system:

\vspace{\baselineskip}
- Poor (Score: 0): The image is repulsive or offensive, lacking any attractiveness. It is completely irrelevant to
the text information and presents the text in a manner that is unpleasant or unacceptable to the audience.

\vspace{\baselineskip}
- Below Average (Score: 1): The image has almost no attractiveness, and the audience is indifferent to it. Its
relevance to the text information is low, and the presentation style is not attractive enough, making the audience
find it rather dull.

\vspace{\baselineskip}
- Average (Score: 2): The image lacks attractiveness, is ordinary and lacks visual highlights, the conveyed text
information is not sufficiently clear, and the presentation style is rather ordinary, lacking novelty or appeal.

\vspace{\baselineskip}
- Above Average (Score: 3): The image's attractiveness is average, without any particular outstanding features
but also not mediocre, conveying the text information and presenting the text in a generally acceptable manner,
albeit not particularly outstanding.

\vspace{\baselineskip}
- Very Good (Score: 4): The image is highly attractive, with good visual effects, conveying the text information
basically, and presenting the text in a way that is appealing to humans, allowing the audience to understand and
resonate to some extent.

\vspace{\baselineskip}
- Excellent (Score: 5): The image is extremely attractive, with outstanding visual effects, clearly and accurately
conveying the text, and presenting the text in a way that resonates deeply with the audience and evokes strong
emotional connections.

\vspace{\baselineskip}
Text: \textcolor{red}{$<$PROMPT$>$}

\vspace{\baselineskip}
Please provide a comprehensive explanation of your score.

\vspace{\baselineskip}
Note that the score has nothing to do with image input order.

\vspace{\baselineskip}
Output format:

\vspace{\baselineskip}
Output for Image 1:

Score: $<$Your Score for Image 1$>$

Explanation: $<$detailed judgment of Score for Image 1$>$

\vspace{\baselineskip}
Output for Image 2:

Score: $<$Your Score for Image 2$>$

Explanation: $<$detailed judgment of Score for Image 2$>$
\end{tcolorbox}

\begin{tcolorbox}[
        title={Aesthetic Judgment Prompt},
        halign=left,
        valign=center,
        nobeforeafter,
    ]
You are a helpful and precise assistant for checking the quality of the answers.

Given the input:

\vspace{\baselineskip}
1. Image 1 and Image 2

2. Question: $\{\{$\textcolor{red}{question}$\}\}$

3. Answer A: $\{\{$\textcolor{red}{answer\_A}$\}\}$

4. Answer B: $\{\{$\textcolor{red}{answer\_B}$\}\}$

\vspace{\baselineskip}
Your task is to evaluate the model's predicted answer, based on the context provided by the images
and the question.
There are two image scores for each answer, and you need to include an evaluation of both outputs
(``Output of Image 1'' and ``Output of Image 2'') in each answer.
Please provide a comprehensive explanation of your score, noting that your explanation should be
based on the facts of the images and not be vague and uninformative.

\vspace{\baselineskip}
Consider the following criteria for evaluation:

\vspace{\baselineskip}
- Relevance: Does each output in the predicted answer relate to the content of each image?

\vspace{\baselineskip}
- Accuracy: Does the prediction in each output accurately reflect the information given in the image
without introducing factual inaccuracies?

\vspace{\baselineskip}
- Objectivity: For the analysis of the images, do the two predicted outputs in each answer give
approximate scores, avoiding any overestimation or underestimation?

\vspace{\baselineskip}
- Clarity: Assess the clarity of the predicted answer. Look for issues such as repetition, unclear
descriptions, or any grammatical errors that could hinder understanding.

\vspace{\baselineskip}
- Completeness: Determine if each predicted output in answer fully covers the scope of the images.
Does it leave out critical information or does it include all necessary details?

\vspace{\baselineskip}
Output Format:

\vspace{\baselineskip}
Output for Answer A:

Score: $<$an integer score of quality from 1-5$>$

Explanation: $<$detailed judgment of prediction for ``Output of Image 1'' and ``Output of Image 2''$>$

\vspace{\baselineskip}
Output for Answer B:

Score: $<$an integer score of quality from 1-5$>$

Explanation: $<$detailed judgment of prediction for ``Output of Image 1'' and ``Output of Image 2''$>$
\end{tcolorbox}


\begin{tcolorbox}[
        title={Pick Judgment Prompt},
        halign=left,
        valign=center,
        nobeforeafter,
    ]
You are a helpful and precise assistant for checking the quality of the answers.

Given the input:

\vspace{\baselineskip}
1. Image 1 and Image 2

2. Prompt: $\{\{$\textcolor{red}{prompt}$\}\}$

3. Question: $\{\{$\textcolor{red}{question}$\}\}$

4. Answer A: $\{\{$\textcolor{red}{answer\_A}$\}\}$

5. Answer B: $\{\{$\textcolor{red}{answer\_B}$\}\}$

\vspace{\baselineskip}
Your task is to evaluate the model's predicted answer, based on the context provided by the images
and the question.
There are two image scores for each answer, and you need to include an evaluation of both outputs
(``Output of Image 1'' and ``Output of Image 2'') in each answer.
Please provide a comprehensive explanation of your score, noting that your explanation should be
based on the facts of the images and not be vague and uninformative.

\vspace{\baselineskip}
Consider the following criteria for evaluation:

\vspace{\baselineskip}
- Relevance: Does each outputs in the predicted answer relate to the content of each image?

\vspace{\baselineskip}
- Accuracy: Does the prediction in each output accurately reflect the information given in the image
without introducing factual inaccuracies?

\vspace{\baselineskip}
- Objectivity: For the analysis of the images, do the two predicted outputs in each answer give
approximate scores, avoiding any overestimation or underestimation?

\vspace{\baselineskip}
- Clarity: Assess the clarity of the predicted answer. Look for issues such as repetition, unclear
descriptions, or any grammatical errors that could hinder understanding.

\vspace{\baselineskip}
- Completeness: Determine if each predicted output in answer fully covers the scope of the images.
Does it leave out critical information or does it include all necessary details?

\vspace{\baselineskip}
Output Format:

\vspace{\baselineskip}
Output for Answer A:

Score: $<$an integer score of quality from 1-5$>$

Explanation: $<$detailed judgment of prediction for ``Output of Image 1'' and ``Output of Image 2''$>$

\vspace{\baselineskip}
Output for Answer B:

Score: $<$an integer score of quality from 1-5$>$

Explanation: $<$detailed judgment of prediction for ``Output of Image 1'' and ``Output of Image 2''$>$
\end{tcolorbox}

\captionof{figure}{\small LVLM-as-a-judge prompt in our model, which enables the model to provide the aesthetic score, pick score and alignment score for each candidate. All scores are based on the rating system, where inputs are assigned scores corresponding to their ratings. The model is fine-tuned in advance with evaluation data to understand the aesthetics of images.}




\end{document}